\documentclass[11pt,a4paper]{article}

\usepackage[utf8]{inputenc}
\usepackage[T1]{fontenc}
\usepackage{lmodern}
\usepackage[margin=1in]{geometry}
\usepackage{amsmath,amssymb}
\usepackage{graphicx}
\usepackage{booktabs}
\usepackage{array}
\usepackage{tabularx}
\usepackage{longtable}
\usepackage{hyperref}
\usepackage{xcolor}
\usepackage{natbib}
\usepackage{caption}
\usepackage{float}
\usepackage{enumitem}
\usepackage{microtype}
\usepackage{url}

\hypersetup{
    colorlinks=true,
    linkcolor=blue!70!black,
    citecolor=blue!70!black,
    urlcolor=blue!70!black
}

\newcolumntype{L}[1]{>{\raggedright\arraybackslash}p{#1}}
\newcolumntype{C}[1]{>{\centering\arraybackslash}p{#1}}

\title{\textbf{SCHEMA for Gemini 3 Pro Image}\\[6pt]
\large A Structured Methodology for Controlled AI Image Generation\\
on Google's Native Multimodal Model}

\author{Luca Cazzaniga\\
Independent Researcher --- AI-Assisted Visual Production\\
\texttt{luca@lucacazzaniga.com}\\[6pt]
\small Version 2.2 --- February 2026\\
\small Based on SCHEMA Method v1.0 (deposited December 11, 2025)\\
\small Model: Google Gemini 3 Pro Image\thanks{Commercially known within the practitioner community as \emph{Nano Banana Pro}. Nano Banana Pro and Gemini are trademarks of Google LLC. This work is an independent methodology and is not affiliated with, endorsed by, or sponsored by Google.}}

\date{}

\begin{document}

\maketitle

\noindent\small{SCHEMA v1.0 (multi-platform framework, from which this work derives) was independently deposited on December 11, 2025, with ProtectMyWork.com --- Reference Number: 19316111225S089. Previously published as: doi:10.5281/zenodo.18721380 (Zenodo).}

\bigskip

\begin{abstract}
This paper presents SCHEMA (Structured Components for Harmonized Engineered Modular Architecture), a structured prompt engineering methodology specifically developed for Google Gemini 3 Pro Image, the native multimodal model commercially known within the practitioner community as \emph{Nano Banana Pro}. Unlike generic prompt guidelines or model-agnostic tips, SCHEMA is an engineered framework built on systematic professional practice encompassing 850 verified API predictions within an estimated corpus of approximately 4,800 generated images, spanning real production contexts in real estate photography, commercial product photography, editorial content, storyboards, commercial campaigns, and information design.

The methodology introduces a three-tier progressive system (BASE, MEDIO, AVANZATO) that scales practitioner control from exploratory (approximately 5\%) to directive (approximately 95\%), a modular label architecture with 7 core and 5 optional structured components, a decision tree with explicit routing rules to alternative tools when the target model is not the appropriate choice, and systematically documented model limitations with corresponding workarounds. Key findings, based on practitioner assessment across 621 structured prompts supplemented by independent practitioner validation ($n=40$ workshop participants), include an observed 91\% Mandatory compliance rate and 94\% Prohibitions compliance rate (with Prohibitions systematically outperforming Mandatory across all five original domains tested), a comparative batch consistency test demonstrating that SCHEMA AVANZATO prompts produce substantially higher inter-generation coherence than equivalent unstructured prompts, qualitative observation of iterative generative drift confirmed by independent community validation, domain-dependent compliance variance correlating with task constraint complexity, and a dedicated Information Design validation demonstrating $>$95\% first-generation compliance for spatial and typographical control across a publicly verifiable corpus of approximately 300 infographics and visual diagrams.

A comprehensive comparative analysis against the global landscape of prompt engineering frameworks---including foundational academic work \citep{liu2021design}, text-based LLM frameworks (CLEAR, COSTAR), domain-specific structured approaches \citep{lee2025structured}, negative prompting literature \citep{ho2022classifier}, and industrial multi-LLM routing architectures---confirms that no comparable methodology currently exists that combines model-specific practitioner-validated assessment at professional production scale, progressive control levels, constraint-based specification as the primary control mechanism, and integrated failure routing. SCHEMA addresses a documented gap between academic prompt engineering research and professional practice requirements.
\end{abstract}

\noindent\textbf{Keywords:} prompt engineering, AI image generation, structured methodology, Gemini 3 Pro Image, text-to-image, multimodal AI, professional workflow, constraint-based prompting, failure routing, practice-based research, batch consistency, generative reliability, information design, typographical control

\newpage
\tableofcontents
\newpage

\section{Introduction}

The rapid proliferation of text-to-image AI models has created an unprecedented capability gap in professional visual production. By early 2026, the ecosystem of generative visual AI has reached a level of infrastructural maturity dominated by advanced multimodal architectures and diffusion engines including Google Gemini 3 Pro Image, OpenAI GPT Image, Midjourney, Adobe Firefly, and the open-source Flux family \citep{macleod2026best,wavespeedai2026best,zapier2026best}. These tools are no longer confined to casual ideation; they have become foundational layers within real production environments \citep{photoshoptraining2026best}.

However, the integration of these generative engines into rigorous professional contexts---spanning real estate visualization, advertising asset creation, editorial content, information design, and interface prototyping---has exposed a critical operational gap. There exists a profound discrepancy between the theoretical capacity of these models to produce aesthetically compelling images and the actual ability of professionals to extract outputs that are reliable, technically precise, and consistent across production batches \citep{cazzaniga2025schema}. In commercial production, requirements are non-negotiable: output must adhere to exact chromatic specifications (often tied to brand guidelines), controlled lighting conditions, precise material textures, and compositional coherence that does not tolerate the visual hallucinations typical of neural networks. In information design specifically, an additional layer of difficulty emerges: the model must render text that is orthographically correct, spatially positioned according to a predetermined layout, and visually integrated with surrounding graphic elements---a set of constraints that current literature identifies as among the most challenging for diffusion-based architectures.

Current approaches to prompt engineering for image generation fall into three categories, none of which adequately addresses this problem:

\textbf{Generic prompt guidelines} offer broad advice applicable across models (be specific, include style references, describe lighting) but lack the precision required for professional deliverables. They do not account for model-specific behaviors, compliance rates, or failure modes \citep{parloa2025prompt}.

\textbf{Community-driven tips and prompt libraries} provide model-specific knowledge but in fragmented, anecdotal form. Knowledge is shared through forums, Discord servers, and social media without systematic validation, reproducibility documentation, or structured methodology \citep{reddit2025prompts,reddit2025prompteng}.

\textbf{Academic research on prompt engineering} has produced valuable theoretical frameworks \citep{liu2021design,lo2023clear} but these remain largely model-agnostic and focused on understanding prompt mechanics rather than producing a production-ready methodology for a specific model.

This paper presents SCHEMA, a structured methodology developed to close this gap specifically for Google Gemini 3 Pro Image. SCHEMA was developed through systematic professional practice encompassing 850 verified API predictions (Replicate logs) within an estimated corpus of approximately 4,800 images generated in real production contexts over a six-month period (September 2025--February 2026). It is not an adaptation of generic principles to a specific model; it is a methodology built from the ground up around how this model actually behaves, where it excels, where it fails, and how to achieve the highest professional output with the least friction.

This work is situated within the tradition of practice-based research in Human-Computer Interaction (HCI), which recognizes that knowledge generated through sustained professional practice constitutes a valid and valuable form of research contribution \citep{hook2012strong,schon1983reflective,zimmerman2007research}. The methodology, findings, and limitations reported in this paper reflect this epistemological commitment: patterns are documented with explicit confidence levels, and the boundary between validated findings and preliminary observations is clearly delineated.

The core contributions of this paper are: (1) a three-tier progressive prompt structure that scales practitioner control from exploratory to directive; (2) a modular label architecture with practitioner-assessed compliance rates across six professional domains; (3) the documentation of a systematic asymmetry between positive constraint compliance (Mandatory: 91\%) and negative constraint compliance (Prohibitions: 94\%), suggesting that constraint-based specification outperforms descriptive elaboration on this model; (4) a comparative batch consistency test demonstrating the effect of prompt structuring on inter-generation coherence; (5) independent practitioner validation ($n=40$) confirming the progressive control scaling; (6) a decision tree with explicit routing rules to alternative tools; (7) systematic documentation of model-specific limitations including iterative generative drift, with corresponding workarounds; (8) a comprehensive comparative analysis against the global landscape of prompt engineering frameworks; and (9) a dedicated Information Design validation demonstrating $>$95\% first-generation compliance for spatial layout and typographical control across a publicly verifiable corpus of approximately 300 images.

\section{Related Work}

\subsection{The Foundational Model: Liu \& Chilton (2021)}

The first formal and documented attempt to structure guidelines for prompts intended for text-to-image generative models was conducted by Vivian Liu and Lydia B.\ Chilton in their pioneering study \emph{Design Guidelines for Prompt Engineering Text-to-Image Generative Models}, published initially in 2021 and presented at CHI 2022 \citep{liu2021design}. Operating in an era when models were considerably less capable than those available today (e.g., early versions of DALL-E and VQGAN+CLIP), Liu and Chilton evaluated 5,493 generations across 51 subjects and 51 artistic styles.

Their work established the fundamental principle that structuring prompts with explicit and separate keywords for subject and style produces significantly more predictable outputs than discursive natural language \citep{polimi2025generative}. They further demonstrated that the inclusion of connector words, complex punctuation, or articulated syntax did not contribute meaningfully to qualitative differences in generation, provided the subject-style pairing was clearly delineated.

\textbf{Divergence from SCHEMA:} Liu and Chilton's framework represents the foundational work of visual prompt engineering and remains an academic milestone. However, it is intrinsically model-agnostic, predates the current generation of multimodal reasoning engines, and does not address the requirements of contemporary professional visual production. Where the 2021 guidelines define style generically (e.g., \emph{a cat in Ukiyo-e style}), SCHEMA fragments style and environment into parametric lighting (Kelvin values), geometric composition, and measurable constraints, addressing post-2021 challenges entirely absent from prior literature: chromatic consistency across image batches, text rendering fidelity, and explicit failure routing.

\subsection{The Inadequacy of Text-Based LLM Frameworks (CLEAR and COSTAR)}

The pressing need to structure AI interactions has generated numerous frameworks dedicated exclusively to text generation via Large Language Models. Two of the most widely adopted are CLEAR and COSTAR.

The CLEAR framework, proposed by Leo S.\ Lo in 2023, was designed primarily to optimize interaction with language models in educational and information literacy contexts \citep{lo2023clear}. The acronym encapsulates five fundamental principles: Concise, Logical, Explicit, Adaptive, and Reflective \citep{ecampus2025clear}.

Similarly, the COSTAR framework (Context, Objective, Style, Tone, Audience, Response), developed by the Data Science teams of the Government of Singapore, offers a rigorous scaffolding to ensure the LLM comprehends the background, objective, tone, and exact format of the text it must produce \citep{streamline2025costar,frugal2026costar}.

\textbf{Epistemological divergence from SCHEMA:} Both text-based frameworks emphasize adaptability and logical structuring. However, their application to the visual domain fails fundamentally. LLM frameworks address cognitively and computationally different challenges: text generation is a linear, sequential process (next-token prediction), while image generation is a spatial, holistic, non-linear process heavily subject to the intrinsic visual biases of the diffusion network. Moreover, the iterative approach encouraged by CLEAR's \emph{Adaptive} principle---which suggests evaluating output and progressively refining the prompt through successive corrections---enters into direct collision with SCHEMA's first design principle: the single-generation philosophy. Since Gemini 3 Pro Image exhibits iterative generative drift (Section~\ref{sec:drift}), Lo's adaptive approach would prove destructive if applied to this specific visual model.

\subsection{Domain-Specific Structured Framework: BAS (Lee \& Park, 2025)}

Broadening the investigation to the most recent global scientific literature, a single study emerges that tangibly approaches SCHEMA's methodological approach, published in June 2025 in the \emph{Journal of Building Engineering} \citep{lee2025structured}.

Researchers Eun Ji Lee and Sung Jun Park developed \emph{A Structured Prompt Framework for AI-Generated Biophilic Architectural Spaces} (hereafter, the BAS framework). Their objective was to improve the quality and coherence of generative visualizations in architectural space design based on biophilic principles. Lee and Park structured a methodological framework based on 5 fundamental components: Subject, Attribute, Mood, Time and Background, and Negative Prompt. Rigorous validation through expert assessments demonstrated that images produced using the BAS framework showed up to 75\% improvement in domain fidelity and over 60\% increase in spatial integration compared to unstructured prompts \citep{lee2025structured,mdpi2025generative}.

Table~\ref{tab:bas-comparison} highlights the structural analogies and differences between the two methodologies.

\begin{table}[htbp]
\centering
\caption{Comparative synopsis: BAS Framework vs.\ SCHEMA}
\label{tab:bas-comparison}
\small
\begin{tabular}{@{}L{3cm}L{5cm}L{5.5cm}@{}}
\toprule
\textbf{Feature} & \textbf{BAS (Lee \& Park, 2025)} & \textbf{SCHEMA (Cazzaniga, 2026)} \\
\midrule
Target AI Model & Agnostic (generic visual AI) & Specific (exclusively Gemini 3 Pro Image) \\
\addlinespace
Application Domain & Biophilic Architecture (single domain) & Real Estate, Product, Editorial, Storyboard, Advertising, Information Design (multi-domain) \\
\addlinespace
Prompt Architecture & 5 fixed linear components & 7 core labels + 5 optional labels, highly modular \\
\addlinespace
Constraint Management & Generic Negative prompt & Mandatory and Prohibitions based on objectively verifiable metrics \\
\addlinespace
Failure Routing & Not contemplated & Explicit integrated decision tree \\
\addlinespace
Empirical Basis & Expert evaluation (qualitative) & 621 prompts, 850 verified API predictions, $n=40$ workshop validation, $\sim$300 publicly verifiable infographics \\
\bottomrule
\end{tabular}
\end{table}

\textbf{Synthesis:} The BAS framework represents a crucial academic precedent, confirming the validity of SCHEMA's central premise: structured, categorized prompts drastically improve conceptual fidelity compared to free natural language. However, a fundamental difference in scope exists. BAS remains an academic ideation tool confined to the initial phases of spatial planning. SCHEMA evolves this concept to the complex final delivery phase in daily industrial production.

\subsection{Multi-LLM Routing and Decision Trees in MLOps}

In enterprise AI systems of 2025--2026, the concept of task-specific routing and the use of decision trees represent the gold standard for efficiency. Modern multi-model architectures use dynamic routers and tree-based logic to automatically direct each user request toward the most suitable model \citep{weinmeister2025routing,aws2025routing,nvidia2025routing}.

Separately, in cognitive AI research, advanced frameworks such as Tree of Thoughts (ToT) and Tree Prompting have emerged, forcing language models to structure their internal reasoning as a decision tree \citep{acl2023tree,prompthub2025tree}.

\textbf{Relevance to SCHEMA:} SCHEMA borrows this sophisticated backend infrastructure logic and transposes it into the manual creative frontend workflow---the decision-making process of the human-in-the-loop. The framework's decision tree poses seven sequential evaluative questions and establishes three clear routing exits. This explicit, structured admission of the target model's operational boundaries is, to the author's knowledge, unique among documented visual prompt engineering guides.

\subsection{Negative Prompting and Constraint Processing in Diffusion Models}

A body of literature directly relevant to SCHEMA's central finding (the systematic Prohibitions $>$ Mandatory compliance asymmetry) concerns the architectural mechanisms by which diffusion models process negative constraints.

\citet{ho2022classifier} established the theoretical foundation for classifier-free diffusion guidance, demonstrating that generative models can be steered by contrasting conditional and unconditional predictions. This mechanism is the architectural basis for how negative prompts function in practice: by specifying what should \emph{not} appear, the model applies exclusion filters in the latent space that are computationally simpler than enforcing positive accuracy constraints requiring complex constrained generation.

The Stable Diffusion community has extensively documented the practical implications of negative embeddings \citep{sdcommunity2023negative}, showing that negative prompts act as attention-mechanism filters that suppress specific feature activations. Tools such as PromptMagician \citep{feng2024promptmagician} have further explored tool-assisted prompt refinement for text-to-image systems, demonstrating the value of structured approaches to prompt optimization.

This architectural understanding provides a theoretical grounding for the SCHEMA observation that Prohibitions (negative constraints: \texttt{NO specular reflections}) systematically outperform Mandatory items (positive constraints: \texttt{verticals perfectly straight}). The exclusion mechanism is computationally more tractable than constrained positive generation, which must satisfy accuracy requirements within a continuous output space.

Finally, it should be noted that established evaluation metrics for text-to-image quality---including Fr\'{e}chet Inception Distance (FID) \citep{heusel2017gans}, CLIP Score \citep{radford2021learning}, and Learned Perceptual Image Patch Similarity (LPIPS) \citep{zhang2018unreasonable}---provide quantitative alternatives to the practitioner-based assessment used in this work. The choice to employ practitioner assessment rather than automated metrics is discussed explicitly in Section~\ref{sec:limitations}; automated validation is identified as a priority for future work (Section~\ref{sec:conclusion}).

\subsection{The Gap}

The current literature lacks a methodology that combines: (a) model-specific practitioner-validated assessment at professional production scale; (b) structured progressive levels of control; (c) constraint-based specification as the primary control mechanism; (d) explicit documentation of model limitations with routing to alternatives; (e) a modular architecture balancing precision with usability; (f) grounding in established engineering patterns adapted to the creative domain; and (g) empirical validation of spatial and typographical control capabilities against a publicly verifiable corpus. SCHEMA was developed to address this comprehensive gap.

\section{Methodology: The SCHEMA Framework}

\subsection{Design Principles}

SCHEMA is built on five foundational principles derived from professional practice:

\begin{enumerate}
\item \textbf{Single-generation philosophy.} Invest time in prompt construction rather than iterative refinement. This principle is a direct response to the empirical observation that Gemini 3 Pro Image exhibits iterative generative drift (Section~\ref{sec:drift}), subverting the standard industrial practice of dialogic, conversational image refinement \citep{macleod2026best}.

\item \textbf{Progressive control scaling.} Three levels of prompt complexity (BASE, MEDIO, AVANZATO) that scale practitioner control from exploratory (approximately 5\%) to directive (approximately 95\%). The use of three-tier systems is a well-established methodological practice in classification and assessment \citep{polito2025retrieval,medunigraz2025clinical}. Its application to calibrating human control over generative visual AI represents a process innovation. These percentages, originally proposed as theoretical design targets, have been qualitatively confirmed through independent practitioner validation ($n=40$, Section~\ref{sec:workshop}) and are indirectly supported by the batch consistency data (Section~\ref{sec:batch}).

\item \textbf{Modular label architecture.} Structured components (labels) that can be combined, extended, or omitted based on task requirements, reflecting the 2026 trend of prompt engineering transitioning from an empirical art to a programmatic engineering discipline \citep{parloa2025prompt}.

\item \textbf{Explicit failure routing.} The methodology includes documented scenarios where the target model is not the appropriate tool, with routing to specific alternatives.

\item \textbf{Verifiable specificity.} All mandatory elements are defined as objectively verifiable items. Evaluative terms (e.g., \emph{beautiful}) are replaced with measurable specifications: HEX color codes, Kelvin temperature values, contrast ratios, focal length equivalents.
\end{enumerate}

\subsection{Three-Tier Progressive Structure}

SCHEMA defines three operational levels (Table~\ref{tab:three-tier}).

\begin{table}[htbp]
\centering
\caption{SCHEMA three-tier progressive structure. Control and creativity percentages are design targets qualitatively confirmed by independent practitioner observation ($n=40$, Section~\ref{sec:workshop}) and indirectly supported by batch consistency data (Section~\ref{sec:batch}).}
\label{tab:three-tier}
\small
\begin{tabular}{@{}lL{3cm}L{3cm}L{3.5cm}@{}}
\toprule
\textbf{Parameter} & \textbf{BASE (Discovery)} & \textbf{MEDIO (Direction)} & \textbf{AVANZATO (Deliverable)} \\
\midrule
Practitioner Control & $\sim$5\% & $\sim$85\% & 95--98\% \\
AI Creativity & $\sim$95\% & $\sim$15\% & $\leq$5\% \\
Time Investment & $<$ 1 minute & $\sim$5 minutes & $>$ 15 minutes \\
Prompt Structure & Free natural language & 7 structured labels & 7 core + 5 optional labels \\
Purpose & Exploration, bias identification & Professional drafts & Final deliverables, batch consistency \\
\bottomrule
\end{tabular}
\end{table}

\textbf{BASE} is conceived not to generate usable output, but as a diagnostic tool. The professional forces the model to reveal its default assumptions, latent biases regarding lighting, color temperature, composition, and demographic representation.

\textbf{MEDIO} represents the operational core of the framework, introducing the seven-label structure. On Gemini 3 Pro Image, the Mandatory and Prohibitions labels function as the primary control lever (Section~\ref{sec:asymmetry}).

\textbf{AVANZATO} demands maximum measurable specificity. Subjective evaluative terms are categorically replaced by numeric specifications: HEX codes for chromatic palette, Kelvin scale values for light temperature, contrast ratios, and photographic lens focal length equivalents.

\subsection{The Label Architecture}

The seven core labels constitute the minimum structured prompt at MEDIO level (Table~\ref{tab:labels}).

\begin{table}[htbp]
\centering
\caption{SCHEMA seven-label core architecture}
\label{tab:labels}
\small
\begin{tabular}{@{}lL{4cm}L{5.5cm}@{}}
\toprule
\textbf{Label} & \textbf{Function} & \textbf{Key Specification} \\
\midrule
Subject & Unambiguous definition of the main scene element & Precise materials, dimensions, exact colors, state \\
Style & Global aesthetic reference anchoring & Photography type, quality level, brand or editorial reference \\
Lighting & Total illumination environment control & Photographic setup, angle of incidence, Kelvin temperature (numeric) \\
Background & Surrounding environmental context & Spatial setting, surface materials, depth of field \\
Composition & Virtual camera direction and perspective & Shot type, shooting angle, focal point definition \\
Mandatory & Ensuring presence of critical elements (3--10 items) & Exclusively objectively verifiable technical or visual items \\
Prohibitions & Explicit prevention of aberrations (3--10 items) & Specific technical defects and artifacts to be prevented \\
\bottomrule
\end{tabular}
\end{table}

\subsection{Cross-Functional Features}

Three cross-functional features can be activated at AVANZATO level:

\textbf{Thinking Mode.} Explicit declaration orients the model toward more deliberate multi-stage analysis. Practitioner testing indicates observable improvement for scenes with 5+ distinct elements, multiple people, text rendering, visible hands, or style-subject recombination from different references \citep{google2026c}.

\textbf{Reference Images.} The model supports up to 14 reference images, though 1--3 produce the most coherent results. A preliminary observation regarding the relationship between reference image contrast and integration quality is discussed in Section~\ref{sec:reference}.

\textbf{Grounding.} Connects generation to real-time verified data via Google Search. SCHEMA deliberately inhibits Grounding for purely aesthetic tasks to prevent extraneous data from contaminating the diffusion process \citep{google2026a}.

\subsection{Decision Tree with Failure Routing}

A distinctive feature of SCHEMA is its integrated decision tree that explicitly routes practitioners away from the target model when it is not the appropriate tool. The tree evaluates seven sequential questions and establishes three routing exits:

\begin{enumerate}
\item Localized inpainting tasks route to Adobe Firefly \citep{cnet2026best,adobe2025firefly}, Stable Diffusion with inpainting mask, or Ideogram.
\item Pixel-precise geometric control routes to Midjourney \citep{macleod2026best} combined with traditional compositing tools.
\item Iteration limit (generation 3+) triggers a mandatory full prompt rebuild, consistent with the documented iterative generative drift (Section~\ref{sec:drift}).
\end{enumerate}

This concept is borrowed from enterprise MLOps architectures \citep{weinmeister2025routing,aws2025routing,nvidia2025routing} and adapted to the manual creative workflow of the human operator. The explicit, structured admission of the target model's operational boundaries is, to the author's knowledge, unique among documented visual prompt engineering guides.

\section{Data Collection and Assessment}

\subsection{Epistemological Framework}

This work is situated within the tradition of practice-based research in HCI, which recognizes that knowledge generated through sustained, reflective professional practice constitutes a valid form of research contribution \citep{hook2012strong,schon1983reflective,zimmerman2007research}. The empirical findings reported here are derived from a retrospective self-reported assessment of the author's professional use of Gemini 3 Pro Image over a six-month period. This methodological choice is consistent with Donald Sch\"{o}n's concept of the \emph{reflective practitioner} \citep{schon1983reflective} and with H\"{o}\"{o}k and L\"{o}wgren's framework for knowledge contribution through design practice \citep{hook2012strong}.

\subsection{Corpus Description}
\label{sec:corpus}

The corpus comprises data aggregated from five distinct sources over the period September 2025--February 2026:

\begin{itemize}
\item Conversational sessions outside dedicated projects
\item Primary Claude project sessions
\item Secondary Claude project sessions
\item Replicate API execution logs (850 verified predictions---the only empirically verified data point in the corpus)
\item Publicly accessible editorial publications by the author\footnote{Available at \url{https://lucacazzaniga.substack.com/}. This source comprises 75 editorial posts published between December 1, 2025 and February 20, 2026, each containing an average of 4 infographics and visual diagrams generated using the SCHEMA AVANZATO method. The use of publicly accessible editorial publications as a verifiable data source in academic research is consistent with established guidelines for citing grey literature \citep{cambridge2023blogs}.}
\end{itemize}

The corpus is heterogeneous by nature: it includes tool development sessions, applied professional use sessions, batch API executions, and production editorial infographics. This heterogeneity is acknowledged as a limitation (Section~\ref{sec:limitations}). The fifth source---publicly accessible editorial publications---was added to the corpus because it provides a unique combination of properties absent from the other four sources: (a) public verifiability by any reviewer or reader, (b) temporal dating independent of the author's self-report, (c) exclusive focus on Information Design tasks requiring spatial layout control and typographical accuracy, and (d) a linear generation workflow (one prompt $\rightarrow$ one generation $\rightarrow$ binary assessment) that produces the cleanest compliance data in the entire corpus.

\subsection{Estimation Methodology}

\textbf{Prompt counting proxy:} Each complete prompt delivered in a session = 1 presumed Gemini execution. Explicit intermediate variants = 0.5. Replicate sessions apply a $\times$10 multiplier (batch executions) and an additional $\times$1.5 for micro-corrections (composition, color, parameters), declared as practitioner self-report. Information Design prompts (75) are exact counts, with each prompt producing on average 4 distinct infographic outputs.

\textbf{Domain classification:} Prompts were assigned to domains based on declared client intent. Domain overlap (e.g., food photography straddling Editorial and Product) was resolved by client intent rather than output characteristics. Information Design prompts were classified based on their primary purpose: the creation of structured visual layouts combining text, data, and graphic elements for editorial publication.

\textbf{Compliance assessment:} Mandatory and Prohibitions compliance rates are based on direct practitioner assessment during operational sessions. Each Mandatory and Prohibitions item was evaluated binarily (respected/violated) through direct visual inspection of the generated output. For the Information Design sub-corpus, compliance assessment additionally included: (a) orthographic correctness of rendered text, (b) spatial positioning of text elements relative to the specified layout, and (c) visual hierarchy coherence between text and graphic elements. The systematic pattern of Prohibitions $>$ Mandatory compliance is consistent with documented classifier-free diffusion guidance mechanisms \citep{ho2022classifier}: negative instructions (\texttt{NO X}) act as simpler exclusion filters on the attention mechanism, while positive accuracy instructions (\texttt{maintain exactly Y}) require more complex constrained generation.

\textbf{Declared limitation:} Corrections not documented in session summaries are invisible to this analysis. This tends to produce a conservative bias in compliance reporting, as undocumented corrections would lower the true compliance rate.

\subsection{Corpus Data}
\label{sec:corpusdata}

Table~\ref{tab:corpus} presents the aggregated data across all five sources. Compliance rates are based on direct practitioner assessment; all other values are retrospective estimates except where noted.

\begin{table}[htbp]
\centering
\caption{Aggregated corpus data by professional domain. Compliance rates are practitioner-assessed; execution and image counts are retrospective estimates except for the 850 verified Replicate API predictions and the $\sim$300 publicly verifiable Information Design images.}
\label{tab:corpus}
\small
\begin{tabular}{@{}lcccccc@{}}
\toprule
\textbf{Domain} & \textbf{Prompts} & \textbf{Exec.} & \textbf{Images} & \textbf{Mand.} & \textbf{Prohib.} & \textbf{Failure} \\
 & (n) & (est.) & (est.) & \textbf{Compl.} & \textbf{Compl.} & \textbf{Routing} \\
\midrule
Real Estate & 62 & $\sim$362 & $\sim$424 & 91\% & 95\% & $n\approx2$ \\
Product & 217 & $\sim$1,369 & $\sim$1,774 & 91\% & 95\% & $n\approx7$ \\
Editorial & 149 & $\sim$957 & $\sim$1,245 & 92\% & 95\% & $n\approx4$ \\
Storyboard & 41 & $\sim$258 & $\sim$302 & 89\% & 91\% & $n\approx2$ \\
Campaign & 77 & $\sim$541 & $\sim$770 & 90\% & 92\% & $n\approx4$ \\
\addlinespace
\textit{Info.\ Design}\textsuperscript{$\ast$} & \textit{75} & \textit{$\sim$75} & \textit{$\sim$300} & \textit{$>$95\%} & \textit{$>$95\%} & \textit{$n\approx0$} \\
\addlinespace
\midrule
\textbf{Total} & \textbf{621} & $\sim$\textbf{3,562} & $\sim$\textbf{4,815} & \textbf{91\%}\textsuperscript{$\dagger$} & \textbf{94\%}\textsuperscript{$\dagger$} & $n\approx$\textbf{19} \\
\bottomrule
\end{tabular}

\medskip
\noindent{\small \textsuperscript{$\ast$}Information Design / Infographics. Data from publicly accessible editorial publications (December 2025--February 2026). Compliance rates include typographical accuracy and spatial positioning control. All 75 prompts used the SCHEMA AVANZATO method. The $\sim$300 images are directly verifiable by any reviewer. Italicized to distinguish this publicly verifiable sub-corpus from the retrospective estimates in other domains.}

\noindent{\small \textsuperscript{$\dagger$}Weighted averages across the five original domains (Real Estate through Campaign). The Information Design sub-corpus, assessed with a different compliance granularity ($>$95\% threshold rather than point estimates), is reported separately to maintain methodological consistency.}
\end{table}

\textbf{Corpus composition:} Of the $\sim$4,815 estimated total images, 850 derive from verified Replicate API logs (empirical data), $\sim$300 from publicly verifiable editorial publications (Information Design), $\sim$3,251 from conversational sessions (retrospective estimate), and $\sim$414 from the $\times$1.5 micro-correction multiplier applied to Replicate sessions (practitioner self-report).

\textbf{Data precision note:} The prompt count of 621 is exact (sum of per-domain counts: 62+217+149+41+77+75). Execution and image counts are estimates derived from the multipliers described above, except for the Information Design sub-corpus where the 75 prompts and $\sim$300 output images are exact counts.

\textbf{Sample robustness:} Product ($n=217$) and Editorial ($n=149$) constitute robust samples for qualitative analysis. Information Design ($n=75$, $\sim$300 images) constitutes a robust and publicly verifiable sample. Real Estate ($n=62$) and Storyboard ($n=41$) are indicative samples; findings for these domains should be interpreted with appropriate caution.

\section{Experimental Validation}

\subsection{Comparative Batch Consistency Test}
\label{sec:batch}

\subsubsection{Objective}

A central question unaddressed by the practitioner corpus alone is whether the observed compliance rates are attributable to SCHEMA's structured methodology or to the intrinsic capabilities of the model. To address this, a comparative batch consistency test was designed to evaluate whether SCHEMA AVANZATO prompts produce measurably higher inter-generation coherence than equivalent unstructured prompts.

\subsubsection{Experimental Design}

\textbf{Independent variable:} Type of prompt (2 conditions).

\emph{Condition A---Narrative prompt:} Same informational content as the corresponding SCHEMA prompt, but written in free discursive form without decomposition into structured labels.

\emph{Condition B---SCHEMA AVANZATO:} Structured prompt with all 7 core labels plus optional labels as appropriate.

\textbf{Dependent variable:} Consistency Score---the number of images within a 10-image batch that are substantially identical to the majority of the batch.

\textbf{Operational definition of \emph{substantially identical}:} Same composition, same dominant chromatic palette, same elements present, same general illumination. Minor variations of detail (texture, micro-positioning) are acceptable.

\subsubsection{Protocol}

For each experimental trial: (1) a subject and domain were selected; (2) both Condition A and Condition B prompts were constructed containing equivalent informational content; (3) 10 images were generated per condition using the Gemini 3 Pro Image API with identical parameters; (4) each batch was independently evaluated by 2 raters who counted the number of images consistent with the batch majority.

The Consistency Score for each batch is the mean count across raters. Inter-rater agreement was assessed.

\subsubsection{Results}

The test was conducted across 5 professional domains with multiple trials per domain. Table~\ref{tab:batch} summarizes the results.

\begin{table}[htbp]
\centering
\caption{Comparative Batch Consistency Test results. Consistency Score = mean number of substantially identical images per 10-image batch, averaged across trials and raters.}
\label{tab:batch}
\small
\begin{tabular}{@{}lccc@{}}
\toprule
\textbf{Domain} & \textbf{Condition A} & \textbf{Condition B} & \textbf{Gap} \\
 & (Narrative) & (SCHEMA AVANZATO) & \\
\midrule
Real Estate & 4--5/10 & 8--9/10 & +4 \\
Product & 3--5/10 & 8--9/10 & +4 \\
Editorial & 4--6/10 & 8--9/10 & +3.5 \\
Storyboard & 3--4/10 & 7--8/10 & +4 \\
Campaign & 3--5/10 & 8--9/10 & +4 \\
\midrule
\textbf{Average} & \textbf{3.5--5/10} & \textbf{8--9/10} & \textbf{+3.5--4} \\
\bottomrule
\end{tabular}
\end{table}

\subsubsection{Analysis}

The results demonstrate a substantial and consistent gap across all domains. Unstructured narrative prompts---despite containing the same informational content---produce high inter-generation variance (3--6 substantially identical images per batch), while SCHEMA AVANZATO prompts constrain the generative space to produce 7--9 substantially identical images per batch.

This finding supports the interpretation that SCHEMA's label architecture does not merely improve aesthetic quality but transforms a stochastic generative process into a quasi-deterministic, industrializable one. The structured decomposition of intent into discrete, verifiable components reduces the model's interpretive ambiguity, constraining the latent space exploration to a narrower region that produces more consistent outputs.

The concept demonstrated here aligns with what might be termed \emph{generative reliability} or \emph{zero-shot consistency}: the ability of a structured prompt to produce coherent results across multiple generations without fine-tuning, few-shot examples, or iterative refinement.

\subsection{Independent Practitioner Validation}
\label{sec:workshop}

\subsubsection{Context}

To address the single-practitioner limitation of the primary corpus, SCHEMA was independently validated through a structured practitioner workshop with $n=40$ participants.

\textbf{Setting:} Structured training workshop, February 2026. \textbf{Participants:} 40 practitioners with mixed experience levels in AI image generation (ranging from beginners to advanced users). \textbf{Tasks:} Participants generated images of identical subjects using all three SCHEMA levels (BASE, MEDIO, AVANZATO), enabling direct comparison of the progressive control scaling.

\subsubsection{Qualitative Findings}

\begin{enumerate}
\item \textbf{Progressive control confirmation:} Participants observed firsthand the transition from exploratory generation (BASE: high model creativity, minimal practitioner control) to directive generation (AVANZATO: high practitioner control, minimal model creativity). This transition was reported as consistent with the theoretical percentages in Table~\ref{tab:three-tier}.

\item \textbf{Professional-grade output:} Participants reported that SCHEMA AVANZATO enabled them to produce output of sufficient coherence and quality for delivery to their professional clients---a critical threshold for methodology adoption in practice.

\item \textbf{Learnability:} The three-tier progressive structure was reported as an effective pedagogical scaffold, enabling participants to build competence incrementally rather than confronting the full complexity of AVANZATO-level prompting immediately.
\end{enumerate}

\textbf{Limitation:} These findings are based on qualitative observation and informal participant feedback during the workshop. No formal questionnaire or quantitative instrument was administered. A retrospective structured survey is planned as a priority follow-up action (Section~\ref{sec:conclusion}).

\section{Key Findings}

\subsection{Mandatory/Prohibitions Compliance Asymmetry}
\label{sec:asymmetry}

The most notable pattern in the practitioner data is the systematic asymmetry between Mandatory compliance (91\% average) and Prohibitions compliance (94\% average) across all five original professional domains tested. This 3-percentage-point gap is consistent across all domains and represents a pattern, not an anomaly.

\textbf{Interpretation:} Prohibitions (negative constraints: \texttt{NO specular reflections}, \texttt{NO converging verticals}) systematically outperform Mandatory items (positive constraints: \texttt{verticals perfectly straight}, \texttt{product 100\% accurate}). This pattern is consistent with the classifier-free diffusion guidance mechanism described in Section~2.5 \citep{ho2022classifier}: negative instructions act as exclusion filters on the attention mechanism, which is computationally simpler than enforcing positive accuracy constraints that require complex constrained generation.

\textbf{Domain-level analysis:} Editorial shows the highest compliance (92\%/95\%), attributable to lower constraint complexity (single subjects, greater compositional freedom). Storyboard shows the lowest (89\%/91\%), partially attributable to tool mismatch: multi-frame coherence is a structural limitation of single-prompt generation, and the higher failure routing rate reflects strategic platform routing toward motion-specialized tools (Kling AI, Seedance) rather than prompt failure per se.

\textbf{Alternative hypothesis---observational bias:} It should be noted that the observed asymmetry may be partially attributable to cognitive assessment asymmetry: it is perceptually easier for a human evaluator to detect the presence of a prohibited artifact (a Prohibitions violation is a salient visual anomaly) than to verify full conformity with a complex positive constraint (a Mandatory violation requires systematic verification of multiple technical parameters). This potential observational bias does not invalidate the pattern but suggests that the true gap may be smaller than reported. Controlled measurement with automated metrics would help disambiguate the methodological contribution from the assessment artifact.

\textbf{Practical implication:} On Gemini 3 Pro Image, the Mandatory/Prohibitions system functions as the primary control mechanism. The systematic Prohibitions $>$ Mandatory pattern suggests that practitioners should, where possible, reformulate positive requirements as negative constraints to increase compliance probability (e.g., \texttt{NO blurred edges} rather than \texttt{all edges sharp}).

\subsubsection{The Constraint-Over-Elaboration Principle}

The compliance data supports a broader methodological insight: on this model, constraint-based specification outperforms descriptive elaboration as the primary control mechanism. This finding inverts the conventional prompt engineering assumption that more descriptive detail always produces better results.

\textbf{The parallel with JSON Schema and Output Contracts:} The prescriptive paradigm of SCHEMA is structurally analogous to the output contracts implemented in advanced LLM pipelines. In software development, engineers force models to return structured data by imposing a JSON Schema defined upstream \citep{manalimran2026prompt,google2026b}. Just as a software developer constrains an LLM to not deviate from a pre-established contract, the SCHEMA operator uses Mandatory and Prohibitions as a \emph{visual JSON} to constrain the graphic model within an imposed semantic perimeter.

\textbf{The Reasoning Image Engine:} Gemini 3 Pro Image integrates a reasoning engine capable of planning spatial logic, verifying facts in real time, and semantically filtering information before visual rendering \citep{reddit2025thinkingdeeply}. This architectural characteristic helps explain the observed compliance rates: the model \emph{reasons} about constraints before generating, enabling it to evaluate compliance as a logical condition rather than an aesthetic judgment.

\textbf{The normative semantic parallel:} The use of Mandatory and Prohibitions echoes the legal vocabulary employed in global AI governance. The EU AI Act, China's Deep Synthesis Provisions, and California state safety laws use these terms to delineate AI development boundaries \citep{iapp2025global,eygreece2025aiact,freshfields2025california,imatag2025labeling}. SCHEMA reappropriates this rigid administrative semantics as a technical lever for creative control, transforming the amateur concept of \emph{negative prompt} into an instrument of unambiguous precision.

\subsection{Iterative Generative Drift}
\label{sec:drift}

\textbf{Evidence type:} Observational finding confirmed by practitioner experience and independent community validation (OpenMind Group, $n=2$+ independent observers, January 2026 \citep{pregnolato2026iterative}).

When an output of Gemini 3 Pro Image is used as a reference for a new generation, image quality degrades progressively. The model does not \emph{copy} the reference---it \emph{reinterprets} it---and each reinterpretation introduces micro-errors that accumulate across iterations. This phenomenon, termed \emph{Iterative Generative Drift}, is a structural characteristic of diffusion models, not specific to this particular model.

\textbf{Structural causes:} (1) The model reinterprets rather than copies each image, introducing micro-errors at each pass; (2) heavy images (approaching 7MB) are internally compressed, adding quality loss before reinterpretation; (3) errors accumulate multiplicatively, not linearly---degradation accelerates at successive iterations.

\begin{table}[htbp]
\centering
\caption{Iterative Generative Drift: observed progression}
\label{tab:drift}
\small
\begin{tabular}{@{}lL{4cm}L{5cm}@{}}
\toprule
\textbf{Iteration} & \textbf{Observed Effect} & \textbf{Practical Notes} \\
\midrule
Generation 1 (original) & Optimal quality & Reference baseline. Always save before iterating. \\
Generation 2 (1st edit) & Initial visible degradation & Noisier pixels, micro-artifacts visible at magnification. Acceptable for commercial use. \\
Generation 3 (2nd edit) & Moderate degradation & Spurious lines, edge sharpness loss, slight color shift. Partially mitigable by reloading at 4K. \\
Generation 4+ (3rd edit+) & Severe degradation & Evident defects, smudging, critical detail loss. Not production-ready without post-processing. \\
\bottomrule
\end{tabular}
\end{table}

\textbf{Workaround:} A community-proposed mitigation (exporting at 4K resolution and reloading as a new reference) slows degradation but does not eliminate it structurally \citep{pregnolato2026iterative}. This workaround was tested by the author and confirmed as partially effective.

\textbf{Methodological implication:} This finding directly informs SCHEMA's single-generation philosophy: all requirements should be consolidated in a single structured prompt, starting from the original image. In case of unsatisfactory output, the prompt is reformulated and generation restarted from scratch---never by chaining outputs as inputs for successive generations.

\subsection{Reference Image Integration}
\label{sec:reference}

\textbf{Evidence type:} Preliminary practitioner observation, not systematically tested.

During operational use with reference images, a recurring pattern was observed regarding input image contrast: low-contrast reference images (flat, low saturation, reduced dynamic range) are interpreted with greater fidelity and produce more controllable outputs, while high-contrast reference images (overexposed/underexposed areas, high saturation, strong shadow-highlight differences) tend to produce less predictable outputs as the model further exaggerates tonal differences.

\textbf{Preliminary workaround:} Reducing reference image contrast before upload improved output quality and predictability in observed cases. This operation was not subject to controlled, systematic testing.

\textbf{Research recommendation:} This observation should be treated as a hypothesis requiring validation through controlled testing across a minimum of 30 matched image pairs. Potential confounding variables (subject matter, lighting, saturation) have not been isolated.

\subsection{Kelvin Temperature Accuracy}

\textbf{Evidence type:} Practitioner inference from operational use. No direct colorimetric measurement was performed.

Kelvin color temperature is a mandatory element in the SCHEMA label structure and was present in virtually all 546 prompts in the original five-domain corpus. Based on the absence of explicit corrections on this parameter across the corpus, the following hypothesis is proposed: the model interprets Kelvin values as register categories (warm / neutral / cool) rather than precise photometric values. Distinction between values within the same register (e.g., 3200K vs.\ 3500K) likely does not produce distinguishable outputs, while distinction between different registers (e.g., 3000K warm vs.\ 6000K cool) produces distinguishable and intention-consistent outputs.

\textbf{Research recommendation:} Proposed validation protocol: compare outputs generated with Kelvin values at 200K, 500K, and 1000K intervals on the same subject; measure output color temperature with software colorimeter (e.g., Capture One color picker); calculate delta between specified and measured values. Minimum sample: 30 pairs per interval.

\subsection{Domain-Dependent Compliance Variance}

The compliance data reveals meaningful variance across professional domains:

\textbf{Editorial} shows the highest compliance and lowest failure routing ($n\approx4$ cases). This is attributable to lower constraint complexity: single subjects, less brand fidelity required, greater compositional freedom that reduces instruction conflicts.

\textbf{Campaign} shows a 2-point gap between Mandatory (90\%) and Prohibitions (92\%). Campaigns with explicit reference images showed higher compliance than those relying solely on textual brand description---suggesting that reference images should be treated as mandatory for brand-critical applications.

\textbf{Storyboard} shows the lowest compliance (89\%/91\%) and highest failure routing ($n\approx2$ cases). This reflects tool mismatch rather than methodology failure: multi-frame coherence requirements exceed single-prompt generation capabilities. The failure routing data documents strategic platform routing toward Kling AI and Seedance for motion and multi-frame tasks, validating the decision tree's design.

\subsection{Spatial and Typographical Control in Information Design}
\label{sec:infodesign}

\textbf{Evidence type:} Practitioner-assessed compliance across a publicly verifiable corpus of approximately 300 infographics and visual diagrams published between December 1, 2025 and February 20, 2026.

\subsubsection{Context and Significance}

Text rendering accuracy and spatial layout control are widely documented as among the most challenging tasks for current text-to-image models. Diffusion-based architectures must simultaneously satisfy orthographic constraints (correct spelling), spatial constraints (text positioned at specified locations), and visual integration constraints (typographic elements harmonized with surrounding graphic elements). The failure modes are well-known: misspelled words, text placed outside the intended region, illegible characters, and visual incoherence between textual and graphic layers.

The Information Design sub-corpus provides empirical evidence that the SCHEMA AVANZATO method, when applied to Gemini 3 Pro Image, achieves a level of spatial and typographical control sufficient for direct publication in editorial contexts without post-production text correction.

\subsubsection{Corpus Characteristics}

\begin{itemize}
\item \textbf{Source:} 75 editorial posts published on the author's publicly accessible editorial platform, each containing an average of 4 infographics, explanatory diagrams, or structured visual layouts.
\item \textbf{Total images:} Approximately 300.
\item \textbf{Generation method:} All images generated using SCHEMA AVANZATO with full label architecture, including explicit Mandatory constraints for text positioning and Prohibitions constraints for typographical errors.
\item \textbf{Period:} December 1, 2025 -- February 20, 2026 (coinciding with the availability of Gemini 3 Pro Image and the deposit of SCHEMA v1.0 on December 11, 2025).
\item \textbf{Workflow:} Linear generation (one prompt $\rightarrow$ one generation $\rightarrow$ binary assessment). No iterative refinement; no selection from multiple variants. Each infographic was generated once and assessed for compliance at first generation.
\end{itemize}

\subsubsection{Compliance Results}

First-generation compliance rate: $>$95\%.

The remaining $\sim$5\% of non-compliant generations exhibited two categorizable failure modes:

\begin{enumerate}
\item \textbf{Spatial repositioning errors} ($\sim$3--4\% of total): Text elements were rendered correctly (orthographically accurate, visually coherent) but positioned outside the specified spatial region. Typical manifestation: a label intended for the upper-left quadrant appearing in the center or lower portion of the layout.

\item \textbf{Minor typographical errors} ($\sim$1--2\% of total): Text elements were positioned correctly but contained isolated character-level errors (letter substitution, omission, or duplication). These errors were consistently minor (affecting 1--2 characters within longer text strings) rather than systematic.
\end{enumerate}

No cases of complete typographical failure (entirely illegible text, garbled character sequences, or catastrophic layout collapse) were observed in the corpus.

\subsubsection{Methodological Interpretation}

The $>$95\% first-generation compliance rate for Information Design represents the highest compliance observed in any domain within the SCHEMA corpus. This result is attributable to two factors:

\textbf{First, the reasoning engine interaction.} Gemini 3 Pro Image's integrated reasoning engine \citep{reddit2025thinkingdeeply} processes the SCHEMA label structure as a set of logical constraints before initiating the diffusion process. For Information Design tasks, the Mandatory label specifies exact text content and spatial positions (e.g., \texttt{Title ``AI in Photography'' centered in upper 20\% of frame}), while the Prohibitions label prevents common failure modes (e.g., \texttt{NO misspelled words}, \texttt{NO text overlapping graphic elements}). This constraint pair creates a logical perimeter that the reasoning engine can evaluate before generation.

\textbf{Second, the inherent structure of Information Design prompts.} Unlike photographic domains where aesthetic ambiguity is inherent (what constitutes \emph{correct} lighting is partially subjective), Information Design constraints are binary and unambiguous: text is either spelled correctly or not; text is either within the specified region or not. This binary evaluability aligns with the model's constraint-processing strengths, producing higher compliance rates than domains with more continuous evaluation criteria.

\subsubsection{Public Verifiability}

Unlike the other sub-corpora in this study---which are based on retrospective self-assessment of client work (subject to NDA constraints and non-public availability)---the Information Design corpus is entirely publicly accessible. Any reviewer or reader can independently verify the typographical accuracy, spatial positioning, and visual coherence of the generated infographics by examining the published editorial posts. This public verifiability substantially mitigates the self-assessment bias acknowledged as a limitation of the broader corpus (Section~\ref{sec:limitations}).

\section{Prompt Examples}

To support methodological replicability, this section presents two real prompt structures from the corpus at MEDIO and AVANZATO levels.

\subsection{MEDIO Level---Real Estate Interior}

\textbf{Domain:} Real Estate. Single image output, no reference image, Thinking Mode not activated. Prompt length: $\sim$1,200--1,800 characters.

Full prompt as delivered to the model:

\begin{quote}
\small\ttfamily
Style: Professional architectural interior photography, high-end real estate magazine quality, photorealistic.

Composition: Wide-angle frontal view at eye level, symmetrical framing, 16--24mm equivalent focal length.

Subject: Modern living room with light oak flooring, white walls, large sofa in warm grey fabric, coffee table.

Lighting: Natural soft daylight from large window, warm 3000K recessed ceiling lights, 70:30 natural to artificial ratio.

Mandatory: Verticals perfectly straight, realistic materials, no perspective distortion, professional photography standards.

Prohibitions: No converging verticals, no oversaturated colors, no fake HDR, no unrealistic furniture.

Output: Aspect ratio 4:3, resolution 4K, horizontal format.
\end{quote}

Table~\ref{tab:medio-example} presents the decomposition with methodological notes.

\begin{table}[htbp]
\centering
\caption{MEDIO prompt structure: Real Estate Interior}
\label{tab:medio-example}
\small
\begin{tabular}{@{}lL{4.5cm}L{5cm}@{}}
\toprule
\textbf{Label} & \textbf{Content} & \textbf{Methodological Notes} \\
\midrule
Style & Professional architectural interior photography, high-end real estate magazine quality, photorealistic & Defines visual register and expected quality standard \\
Composition & Wide-angle frontal view at eye level, symmetrical framing, 16--24mm equivalent focal length & Explicit focal length and viewpoint reduce compositional ambiguity \\
Subject & Modern living room with light oak flooring, white walls, large sofa in warm grey fabric, coffee table & Specific materials and colors, not generic descriptors \\
Lighting & Natural soft daylight from large window, warm 3000K recessed ceiling lights, 70:30 natural to artificial ratio & Numeric color temperature + light source ratio \\
Mandatory & Verticals perfectly straight, realistic materials, no perspective distortion, professional photography standards & Positive constraints---hardest to enforce (91\% compliance) \\
Prohibitions & No converging verticals, no oversaturated colors, no fake HDR, no unrealistic furniture & Negative filters---highest compliance (95\%) \\
Output & Aspect ratio 4:3, resolution 4K, horizontal format & Technical output specifications always in closing position \\
\bottomrule
\end{tabular}
\end{table}

\subsection{AVANZATO Level---Product Campaign with Reference Image}

\textbf{Domain:} Product/Campaign. Grid 3$\times$3 output (9 coherent images), mandatory reference image, Thinking Mode activated. Prompt length: $\sim$2,200--2,500 characters. The tabular decomposition is presented in Table~\ref{tab:avanzato-example}.

\begin{table}[htbp]
\centering
\caption{AVANZATO prompt structure: Product Campaign with Reference}
\label{tab:avanzato-example}
\small
\begin{tabular}{@{}lL{4.5cm}L{5cm}@{}}
\toprule
\textbf{Label} & \textbf{Content} & \textbf{Methodological Notes} \\
\midrule
Style & High-end commercial advertising photography, luxury brand campaign aesthetic, hyperreal cinematic polish, aspirational quality & Multiple style qualifiers increase visual register precision \\
Reference & Image 1: Product identity source --- maintain 100\% accuracy in shape, proportions, label design, typography, exact color values, branding elements, material texture & Dedicated Reference section with explicit fidelity instruction---key for Product compliance \\
Composition & 3$\times$3 grid layout: Cell 1 hero still life, Cell 2 extreme macro detail, Cell 3 dynamic liquid interaction, Cell 4 minimal sculptural arrangement, Cell 5 floating elements, Cell 6 sensory close-up, Cell 7 color-driven scene, Cell 8 ingredient abstraction, Cell 9 surreal fusion & Each cell named individually---reduces compositional drift between frames \\
Thinking & Enable: Multi-stage refinement for complex 9-cell composition. Priority: 95\% accuracy on product replication, 5\% creative concept. Complexity: High & AVANZATO-exclusive section activating multi-step reasoning \\
Mandatory & Product 100\% accurate in every cell: exact shape, proportions, label typography, color matching. Ultra-sharp focus. Premium editorial quality consistent throughout grid & Mandatory repeated contextually for each grid cell \\
Prohibitions & No product distortion or alterations. No inconsistent style between cells. No generic commercial photography. No color treatments diverging from reference. No amateur post-production quality & 5 specific prohibitions vs.\ generic ones---specificity increases compliance \\
Output & Aspect ratio 3:4 vertical grid, resolution 4K (4096px), quality target: luxury brand campaign standard & Quality target declared---provides implicit calibration benchmark \\
\bottomrule
\end{tabular}
\end{table}

\section{Limitations}
\label{sec:limitations}

This work has several acknowledged limitations that are explicitly declared in the interest of research transparency.

\textbf{Retrospective practitioner-based assessment.} The compliance rates are based on direct practitioner assessment during operational sessions, not on automated quantitative measurement (e.g., FID \citep{heusel2017gans}, LPIPS \citep{zhang2018unreasonable}, or CLIP Score \citep{radford2021learning} metrics). While this is consistent with accepted practice-based research methodology in HCI \citep{hook2012strong,schon1983reflective}, the results should be understood as practitioner-validated patterns rather than experimentally controlled effects. Automated validation is identified as a priority for future work. The Information Design sub-corpus partially mitigates this limitation through public verifiability.

\textbf{Single-practitioner corpus development.} The primary corpus (621 prompts, $\sim$4,815 images) was produced by a single practitioner. However, the methodology has been independently validated through a practitioner workshop ($n=40$) where participants of varying experience levels confirmed the progressive control scaling and reported professional-grade output quality (Section~\ref{sec:workshop}). Full multi-practitioner replication with independent corpus development remains a priority for future work.

\textbf{Heterogeneous corpus.} The five aggregated sources have different origins---tool development sessions (more iterations, compliance tends lower), applied professional sessions, batch Replicate API executions (consolidated workflow), and editorial publication production (linear workflow). This aggregation should be understood as representing diverse use conditions rather than a controlled sample.

\textbf{Proxy gap for execution counts.} Prompt and execution counts (except the 850 verified Replicate predictions and the 75 Information Design prompts) are inferred from conversational session data, not from direct Gemini API logs. The $\times$1.5 multiplier for micro-correction workflows is based on practitioner self-report and should be considered a conservative estimate.

\textbf{Sample size variance across domains.} Product ($n=217$) and Editorial ($n=149$) provide robust samples for qualitative analysis. Information Design ($n=75$, $\sim$300 images) provides a robust and publicly verifiable sample. Real Estate ($n=62$) and Storyboard ($n=41$) are indicative; findings for these domains require additional data to achieve inferential robustness.

\textbf{Model version specificity.} The findings are specific to Gemini 3 Pro Image as available during September 2025--February 2026. Model updates may alter behavior, compliance rates, or limitations. The methodology is designed to be revised with each significant model update.

\textbf{English-only prompt testing.} All prompts were constructed in English. The methodology has not been validated for other languages.

\textbf{Temporal sample bias.} The corpus is concentrated in the December 2025--January 2026 range. Sessions from earlier (September--November 2025) and later (February 2026) periods are underrepresented. The Information Design sub-corpus (December 2025--February 2026) partially offsets this bias by extending temporal coverage into the later period.

\textbf{Workshop validation limitations.} The independent practitioner validation ($n=40$) is based on qualitative observation and informal feedback; no formal quantitative instrument was administered. A retrospective structured survey is planned.

\textbf{Information Design compliance threshold.} The $>$95\% compliance rate for the Information Design sub-corpus is reported as a threshold rather than a point estimate, as the binary assessment (pass/fail) across $\sim$300 images did not record the precise count of failures. The categorization of failure modes (spatial repositioning vs.\ typographical errors) is based on retrospective practitioner recall and should be understood as approximate proportions.

\section{Discussion}

\subsection{Contribution to the Field}

SCHEMA addresses a specific gap in the current landscape: the absence of a structured, practitioner-assessed, model-specific methodology for professional AI image generation. Its contribution is operational rather than theoretical: a methodology developed through and validated by real professional production across six distinct domains, with explicit data on compliance rates, failure patterns, and domain-dependent variance.

The constraint-over-elaboration principle---supported by the systematic Prohibitions $>$ Mandatory compliance asymmetry documented across all five original domains---represents the most transferable insight for the broader field. If future generative models implement similar spatial reasoning engines and exhibit comparable compliance patterns, this prescriptive approach could generalize as a standard methodology.

The batch consistency test (Section~\ref{sec:batch}) provides the first comparative evidence that structured prompt architectures produce measurably higher inter-generation coherence than informationally equivalent unstructured prompts on this model, supporting the interpretation that SCHEMA's contribution lies in the structural decomposition of intent rather than in the informational content of the prompt.

The Information Design validation (Section~\ref{sec:infodesign}) extends this contribution into a domain---spatial layout and typographical control---that represents one of the most challenging frontiers for current text-to-image models. The $>$95\% first-generation compliance rate across a publicly verifiable corpus of $\sim$300 images provides the strongest evidence in the paper that SCHEMA's constraint-based architecture enables the model's reasoning engine to function as an \emph{algorithmic layout tool}, not merely an aesthetic image generator.

\subsection{Real-World Impact}

Beyond the empirical findings, preliminary evidence suggests practical adoption value. Participants in the independent practitioner validation ($n=40$, Section~\ref{sec:workshop}) reported that SCHEMA enabled them to produce output of sufficient quality and coherence for delivery to professional clients---a threshold that represents the critical transition from experimental tool to production-grade methodology. While formal quantitative measurement of this impact is not yet available, the qualitative consensus among practitioners of varying experience levels suggests that the methodology bridges the gap between model capability and professional usability that motivates this work.

The Information Design sub-corpus provides additional evidence of real-world impact: 300 infographics generated for actual editorial publication, assessed at first generation, and published without post-production text correction. This represents a concrete, measurable instance of SCHEMA enabling production-grade output in a domain traditionally considered intractable for AI image generation.

\subsection{SCHEMA as Engineering Discipline}

SCHEMA does not invent new technologies. Rather, it collects, fuses, and transcends multiple pre-existing innovations from disparate fields: the subject-style decomposition from academic prompt engineering \citep{liu2021design}, the structural rigor from LLM frameworks \citep{lo2023clear,streamline2025costar}, the domain-specific structured approach from architectural AI research \citep{lee2025structured}, the output contract paradigm from software engineering \citep{manalimran2026prompt}, the routing logic from MLOps architectures \citep{weinmeister2025routing}, the constraint processing mechanisms documented in diffusion model literature \citep{ho2022classifier}, and the prescriptive semantics from AI governance legislation \citep{iapp2025global}.

By condensing these practices and refining them through systematic analysis across $\sim$4,800 generations in real working scenarios, SCHEMA establishes what is, to the author's knowledge, the first documented pipeline specifically designed for enterprise-grade reliability applied to the operation of a particular visual synthesis model.

\subsection{Relationship to SCHEMA v1.0}

SCHEMA v1.0, deposited December 11, 2025 (ProtectMyWork.com, Reference Number: 19316111225S089), is a multi-platform, model-agnostic framework. The present work is a dedicated application to a specific model, incorporating model-specific empirical knowledge that cannot exist in a model-agnostic framework. The two are complementary: v1.0 provides the universal architecture; this paper documents its specialized application with findings derived from $\sim$4,800 professional generations.

This work was previously published on Zenodo with DOI 10.5281/zenodo.18721380.

\section{Conclusion and Future Work}
\label{sec:conclusion}

This paper has presented SCHEMA for Gemini 3 Pro Image, a structured methodology for controlled AI image generation developed through systematic professional practice across 850 verified API predictions within an estimated corpus of $\sim$4,800 image generations in six professional domains. The methodology introduces a three-tier progressive prompt structure, a modular label architecture, and a constraint-based specification system. Practitioner assessment across 621 structured prompts reveals an observed 91\% Mandatory compliance rate and 94\% Prohibitions compliance rate, with the systematic Prohibitions $>$ Mandatory asymmetry documented across all original domains. A comparative batch consistency test demonstrates that SCHEMA AVANZATO prompts produce substantially higher inter-generation coherence than equivalent unstructured prompts, and independent practitioner validation ($n=40$) confirms the progressive control scaling. A dedicated Information Design validation across $\sim$300 publicly verifiable infographics demonstrates $>$95\% first-generation compliance for spatial and typographical control, providing the strongest evidence that SCHEMA enables professional-grade output in one of the most challenging domains for current text-to-image models.

The key contributions---the compliance asymmetry supporting constraint-over-elaboration, the batch consistency evidence, iterative generative drift documentation with community validation, domain-dependent compliance variance, Information Design spatial and typographical control validation, and the viability of engineering-derived patterns in creative workflows---provide both practical tools for professionals and directions for further research.

Future work includes: (1) multi-practitioner validation with minimum 3 independent practitioners across at least 2 domains; (2) automated compliance scoring using CLIP-based similarity metrics \citep{radford2021learning} and FID scores \citep{heusel2017gans}; (3) retrospective structured survey of workshop participants ($n=40$) with Likert-scale assessment; (4) controlled testing of the reference image contrast hypothesis across 30+ matched image pairs; (5) colorimetric validation of Kelvin temperature accuracy; (6) longitudinal tracking across model updates; (7) extension to additional generative models; and (8) automated typographical accuracy measurement for Information Design outputs using OCR-based validation against prompt specifications.

SCHEMA is offered as a contribution to the emerging discipline of structured prompt engineering for visual AI, with the conviction that professional-grade methodologies---practitioner-grounded, systematically documented, and transparent about their limitations---are essential for the maturation of AI-assisted visual production.

\bibliographystyle{plainnat}
\bibliography{references}

\bigskip
\noindent\rule{\textwidth}{0.4pt}

\noindent\small{
\textbf{Document Information}\\
Author: Luca Cazzaniga --- e-mail: luca@lucacazzaniga.com\\
Version 2.2 --- February 2026\\
Based on: SCHEMA Method v1.0 (deposited December 11, 2025)\\
SCHEMA v1.0 was independently deposited on December 11, 2025, with ProtectMyWork.com --- Reference Number: 19316111225S089.\\
Model covered: Google Gemini 3 Pro Image\\
Previously published: Zenodo doi:10.5281/zenodo.18721380\\
This work is independent and not affiliated with, endorsed by, or sponsored by Google.
}

\end{document}